%% file: main.tex
\documentclass{article}
\usepackage{tcolorbox,needspace}
\tcbuselibrary{breakable,skins}
\definecolor{promptblue}{HTML}{536F99}
\definecolor{promptbackground}{HTML}{F2F5FA}
\definecolor{prompttext}{HTML}{26354A}
\newtcolorbox{navprompt}[1]{
  enhanced,
  colback=promptbackground,colframe=promptblue,
  colbacktitle=promptblue!18!white,coltitle=prompttext,
  colupper=prompttext,
  title={#1},fonttitle=\sffamily\bfseries\small,
  fontupper=\fontfamily{fvm}\fontsize{9}{11.5}\selectfont,
  arc=2mm,boxrule=0.6pt,
  left=9pt,right=9pt,top=8pt,bottom=8pt,
  toptitle=5pt,bottomtitle=5pt,
  before skip=9pt,after skip=9pt,
  before upper={\raggedright\setlength{\parskip}{5pt}\setlength{\parindent}{0pt}}
}

\usepackage{iclr2027_conference,times}
\usepackage[T1]{fontenc}
\usepackage{amsmath,amssymb,booktabs,array,tabularx,xcolor,graphicx}
\usepackage{hyperref}
\usepackage{url,xspace,capt-of,placeins,wrapfig}
\usepackage{colortbl}
\usepackage{algorithm,algpseudocode}
\usepackage{booktabs}
\usepackage{array}
\usepackage[table]{xcolor}

\title{NavHarness: Adaptive Goals for\\Vision-and-Language Navigation}

\iclrfinalcopy

\title{%
  \centering
  NavHarness: Adaptive Goals for\\
  Vision-and-Language Navigation
}

\author{%
  \parbox{\dimexpr\textwidth-2\tabcolsep\relax}{%
    \centering
    \normalfont\normalsize
    \textbf{Haoxiang Shi}\textsuperscript{1,2}%
    \quad
    \textbf{Zaijing Li}\textsuperscript{1,2}%
    \quad
    \textbf{Muhe Ding}\textsuperscript{1}%
    \quad
    \textbf{Xiang Deng}\textsuperscript{1}%
    \\[4pt]
    \textbf{Yaowei Wang}\textsuperscript{1,2}%
    \quad
    \textbf{Liqiang Nie}\textsuperscript{1}%
    \\[10pt]
    \textsuperscript{1}Harbin Institute of Technology (Shenzhen)%
    \\[2pt]
    \textsuperscript{2}Pengcheng Laboratory\\[6pt]Project Page: {\hypersetup{hidelinks}\href{https://Navharness.github.io}{\textcolor{blue}{\nolinkurl{https://Navharness.github.io}}}}
    \\[8pt]
  }%
}

\newcommand{\method}{NavHarness\xspace}
\newcommand{\pending}{\textcolor{gray}{\textemdash}}
\newcommand{\up}{\ensuremath{\uparrow}}
\newcommand{\down}{\ensuremath{\downarrow}}

\begin{document}
\raggedbottom
\maketitle
\lhead{Preprint}
\input{sections/abstract}

\input{sections/introduction}
\input{sections/related_work}
\input{sections/method}
% \input{sections/experiments_new}
\input{sections/experiments_revised_v2}
\FloatBarrier
\input{sections/conclusion}

% \clearpage
\bibliography{references}
\bibliographystyle{iclr2027_conference}
\clearpage
\appendix
\input{sections/appendix}
\end{document}

%% file: sections/abstract.tex
\begin{abstract}
Vision-Language Navigation (VLN) requires embodied agents to generate actions based on instructions and observations. General-purpose multimodal agents offer a promising basis for this task, but selecting plausible local actions does not ensure that execution remains consistent with the intended route, particularly in long-horizon tasks. Moreover, the accumulated interaction history increases the input required for subsequent decisions, resulting in a significant inference overhead. To this end, we introduce \method, an Agentic VLN framework that includes a Goal Agent that sets adaptive goals for local actions, a Verify Agent that dynamically verifies whether a goal has been completed, a Memory Agent for multimodal context compression, and a Visuomotor Agent to execute adaptive goals. Specifically, the Goal Agent formulates adaptive goals based on the instruction, current observation, and execution history. Then the Visuomotor Agent executes navigation actions to achieve each goal, while the Verify Agent uses a goal-specific verification question to dynamically assess whether the observed outcomes satisfy the intended completion condition. Verified goal completion then marks a boundary for the Memory Agent to compress the corresponding multimodal interaction history while preserving information needed for subsequent navigation. We evaluate navigation on R2R-CE and RxR-CE, examine framework variants across three model backbones, and study context evolution during execution. For Real-World evaluation, \method achieves 83.3\% success and 1.51\,m navigation error across eight challenging routes evaluated three times each.
\end{abstract}

% \begin{abstract}
% Vision-language navigation requires an agent to keep a route instruction aligned with observations gathered while moving through an unfamiliar environment. General-purpose multimodal agents offer a promising basis for this task, but selecting plausible local actions does not ensure that execution remains consistent with the intended route. This difficulty becomes more pronounced as trajectories lengthen, while accumulated interaction history increases the input required for subsequent decisions. We introduce \method, an agent framework that uses adaptive navigation goals as a common unit for action, progress monitoring, and context management. The framework generates the next local goal from the instruction, current observation, and execution history, allowing navigation to respond to the situation actually encountered. Each goal is paired with a verification question that checks whether observed outcomes satisfy the intended completion condition. Verified completion then provides a boundary for summarizing completed history while retaining information needed for subsequent navigation. We evaluate navigation on R2R-CE and RxR-CE, examine framework variants across three model backbones, and study context evolution during execution. In physical environments, \method achieves 83.3\% success and 1.51\,m navigation error across eight challenging routes evaluated three times each. The framework provides a task-grounded organization for connecting local decisions to sustained instruction following.
% \end{abstract}

%% file: sections/introduction.tex
\begin{figure}[!ht]
\centering
\includegraphics[width=\linewidth]{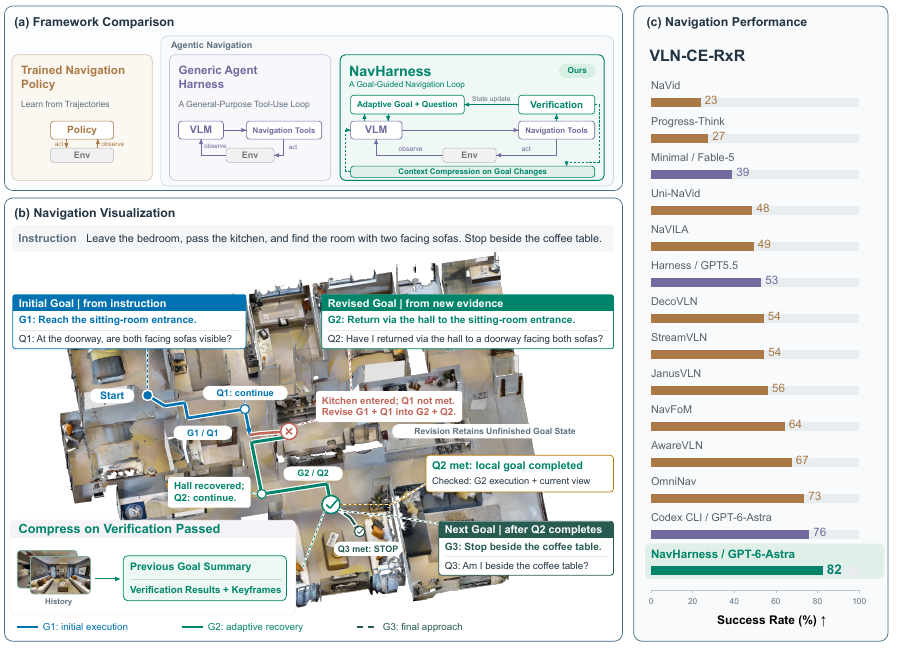}
\caption{
\textbf{NavHarness connects adaptive goal execution, verified progress, and multimodal memory.} Adaptive local goals guide navigation, while goal-specific questions verify whether observed outcomes meet completion conditions. Verified completion marks boundaries for summarizing completed interactions while retaining information for ongoing navigation and recovery. NavHarness significantly outperforms trained policies and other agent frameworks in VLN-CE.}
\label{fig:teaser}
\end{figure}

\section{Introduction}
\label{sec:intro}

Vision-Language Navigation (VLN) requires an embodied agent to follow natural-language instructions using visual observations~\citep{anderson2018vln}. In continuous environments, the agent must translate route descriptions into physical movement while avoiding obstacles and recognizing when it has reached the destination~\citep{krantz2020vlnce}. Training on large-scale expert trajectories has driven substantial advances in task-specific policies and navigation foundation models~\citep{zhang2024navid,cheng2024navila,qwen2026robotnav}. However, an instruction cannot prescribe the appropriate response to every situation encountered during execution. Following the intended route therefore requires an agent to continually interpret new observations in the context of the instruction and adapt its actions as the environment is revealed.

General-purpose multimodal agents offer a promising basis for such adaptive decision-making. Agent frameworks allow models to interleave reasoning, tool use, and environmental feedback~\citep{yao2023react}. In VLN, this interaction pattern enables a multimodal model to interpret the current scene, execute movements through navigation tools, and use the resulting observations to guide subsequent decisions. Recent navigation agents have demonstrated the potential of this approach through competitive performance with navigation foundation models~\citep{li2026agenticnav,zhou2026mip,chen2026harnessvln}. However, deciding what to do at an individual step is only part of the problem. Sustained navigation also requires keeping local execution consistent with the route instruction and managing the information carried forward between decisions.

\begin{samepage}
Two challenges arise in this setting. First, \emph{locally plausible actions do not ensure consistency with the intended route}. Partial observations, detours, and mistaken turns can cause the actual trajectory to diverge from the agent's assumed progress. Without checking whether an intended local objective has been achieved, the agent may base subsequent decisions on an incorrect interpretation of its progress along the route, allowing deviations to compound over long horizons~\citep{zhou2026mip}. Second, \emph{accumulating interaction history increases inference overhead}. Repeatedly including past observations and tool interactions enlarges the input required for subsequent decisions. Indiscriminate truncation, however, can remove landmarks, corrections, or unfinished objectives that remain necessary for navigation and recovery. An explicit distinction between what has been accomplished and what remains unresolved offers a common basis for addressing both challenges.\par
\end{samepage}

To this end, we introduce \method, an agentic VLN framework comprising a Goal Agent, a Verify Agent, a Memory Agent, and a Visuomotor Agent (Figure~\ref{fig:teaser}). To keep local execution aligned with the route instruction, the framework couples \emph{adaptive goal generation with explicit completion verification}. The Goal Agent formulates local goals from the instruction, current observation, and execution history, allowing the immediate objective to reflect the situation actually encountered. Each goal is paired with a verification question that specifies an observable completion condition. At each step, the Visuomotor Agent executes actions toward the goal, while the Verify Agent checks subsequent observations to assess whether that condition has been satisfied. Verification feedback guides whether to continue execution, advance to the next goal, or revise the objective to recover from a deviation while retaining unfinished task state. This separates goal pursuit from completion assessment, grounding progress updates in observed outcomes rather than issued actions alone.

Moreover, we introduce \emph{progress-aware multimodal context compression} to limit the growth of interaction history while preserving information needed for navigation. The Memory Agent uses verified goal completion as a boundary for compression, distinguishing completed interactions that can be summarized from context still needed for ongoing execution. It condenses the multimodal history of completed goals into compact summaries that preserve route progress, relevant landmarks, and corrections, while retaining recent evidence for the active goal and carrying unfinished state forward when goals are revised. Subsequent decisions can thus build on prior progress without repeatedly processing the full sequence of past observations and interactions. By coupling compression boundaries and information retention to verified progress, this design reduces the historical context carried into subsequent decisions while avoiding the loss of unresolved objectives and recovery-relevant evidence that indiscriminate truncation can cause.

We evaluate \method{} on R2R-CE and RxR-CE, compare framework variants across three model backbones, and analyze context evolution during execution. With GPT-6-Astra as backbone, \method{} achieves success rates (SR) of 79.0\% on R2R-CE and 82.5\% on RxR-CE, compared with 74.0\% and 76.3\%, respectively, for Codex CLI using the same backbone. Analyses of execution traces further show that compression reduces decision input at selected navigation stages. In real-world evaluation, \method{} achieves 83.3\% SR and 1.51\,m navigation error across eight challenging routes, each evaluated three times.

Our contributions are threefold:
\begin{itemize}
    \item We introduce \method, an agentic VLN framework that couples adaptive local goals with goal-specific completion verification, grounding progress updates in observed outcomes and supporting continued execution and recovery along the intended route.
    \item We develop a progress-aware multimodal memory mechanism that uses verified goal completion to define compression boundaries, summarizing completed interactions while preserving the context and unfinished task state needed for subsequent navigation.
    \item Experimental results on VLN-CE, and Real-World show that the \method outperforms navigation foundation models which trained on large-scale datasets in a zero-shot setting. 
\end{itemize}

%% file: sections/related_work.tex
\section{Related Work}
\label{sec:related}

\subsection{From navigation policies to agents.}
Navigation models learn instruction grounding and action selection from trajectories~\citep{zhang2024navid,cheng2024navila,qwen2026robotnav}. Language-model systems reason over observations and feedback~\citep{zhou2024navgpt,chen2024mapgpt}, while recent agents control environmental interaction through tool calls~\citep{li2026agenticnav,zhou2026mip,chen2026harnessvln}.  In \method, adaptive goals guide execution, while verified completion governs goal advancement and memory compression. Appendix A provides an extended discussion of the literature.

\subsection{Progress assessment and recovery.}
Progress estimates and backtracking help agents detect and correct deviations~\citep{ma2019selfmonitoring,ma2019regretful}. Recent methods incorporate explicit state reasoning or waypoint-based recovery~\citep{guo2026awarevln,shi2025smartway}. \method makes each adaptive goal's completion condition explicit through a verification question. The resulting judgment guides goal continuation, revision, or advancement and identifies completed interaction for compression.

\subsection{Memory and context management.}
Navigation systems retain temporal context, structured memories, or language summaries~\citep{wei2025streamvln,zeng2026janus,zhou2024navgpt}. Prompt compression and context-utilization studies further motivate selective retention~\citep{pan2024llmlingua2,liu2024lostmiddle}. \method ties compression to verified goal completion, preserving current observations and unresolved route information while summarizing completed interaction.

%% file: sections/method.tex
\section{NavHarness}
\label{sec:method}

\subsection{Task Formulation and Framework Overview}
\label{sec:setup}

\begin{figure*}[!t]
\centering
\includegraphics[width=\textwidth]{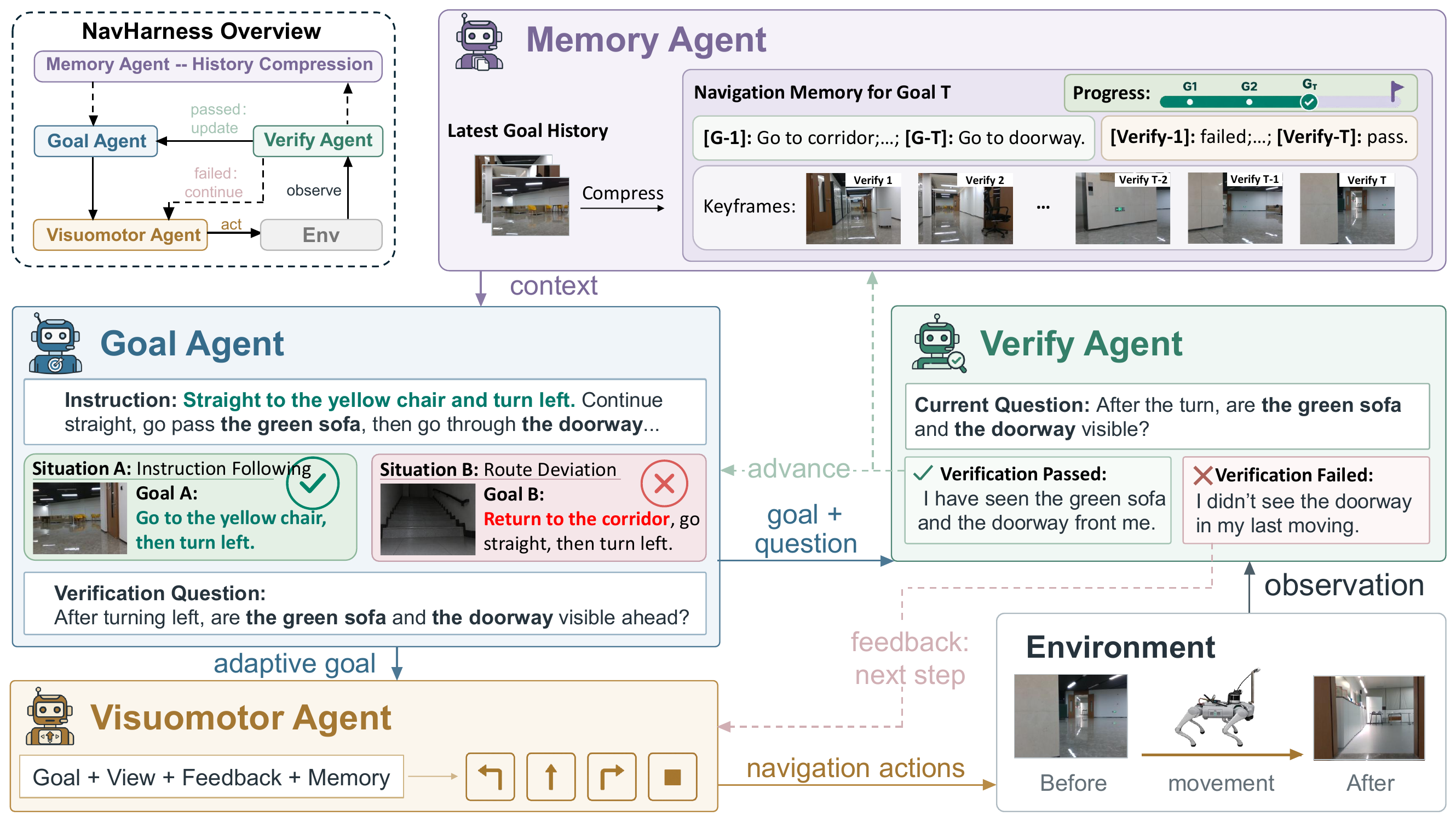}
\caption{\textbf{NavHarness framework.} Adaptive goals guide execution; observation-based verification provides feedback for the next step and goal revision or advancement. Verified completion triggers multimodal memory compression. The two situations illustrate route following and recovery.}
\label{fig:framework}
\end{figure*}

% Role: task setting and available information.
Given an instruction $x$ and an initial RGB observation $o_0$, an embodied agent must navigate to the specified destination in an unfamiliar environment. At interaction step $t$, it selects an action or short action sequence $a_t$ using the instruction, current view, and history. Execution returns observation $o_{t+1}$ and an execution record. Each step may include several low-level movements. 

% Role: framework overview and the connection between progress and memory.
As shown in Figure~\ref{fig:framework}, \method{} organizes navigation around adaptive goals with observable completion conditions. The Goal Agent specifies a local goal and its verification question, the Visuomotor Agent executes actions toward the goal, and the Verify Agent assesses the observed outcomes. Verification feedback guides continued execution, goal revision, or advancement to the next goal. Verified completion also defines a boundary for the Memory Agent to compress the corresponding history. The updated memory supplies context for subsequent goal generation, execution, and verification, linking navigation progress to context management.

% Role: notation and shared context.
We denote the active goal by $g_k$ and its verification question by $q_k$, where $k$ indexes goal updates and $t$ indexes interaction steps. The shared historical context is $c_t=(m_t,r_t)$: $m_t$ contains compressed memory and retained visual evidence, while $r_t$ contains recent interaction records, including available verification feedback. The original instruction and current observation remain available alongside this context.

\subsection{Goal Agent: Adaptive Local Goals}
\label{sec:goals}

% Role: goal generation and observable completion conditions.
The Goal Agent generates one local goal at a time, grounding it in the encountered situation:
\begin{equation}
    (g_k,q_k)=\mathcal{G}(x,o_{\tau_k},c_{\tau_k}),
    \label{eq:goal}
\end{equation}
where $\tau_k$ is the interaction step at which goal $k$ is established. The goal $g_k$ specifies a desired navigation outcome, while $q_k$ identifies the visual evidence or spatial relation that would support its completion. A goal may span multiple interaction steps, allowing the Visuomotor Agent to adapt its movements to the current view while pursuing the same objective.

% Role: goal revision and preservation of unfinished state.
When observations or verification feedback indicate that the active goal no longer fits the scene, the Goal Agent formulates a corrective goal from the updated context. Figure~\ref{fig:framework} illustrates this distinction: a route-following goal directs the agent toward the yellow chair and then to turn left, whereas an unexpected stairway motivates a corrective goal to return to the corridor before resuming the route. The original instruction constrains the revised goal, and unfinished state is carried forward. Goal revision supports recovery without marking an unresolved route stage as complete.

\subsection{Verify Agent: Goal-Specific Completion Verification}
\label{sec:verification}

% Role: verification timing, feedback content, and evidence requirements.
The Verify Agent assesses goal completion after each interaction step, separating outcome assessment from action generation:
\begin{equation}
    p_{t+1}=\mathcal{V}(x,g_k,q_k,o_{t+1},c_{t+1}^{-}).
    \label{eq:verify}
\end{equation}

Here, $c_{t+1}^{-}$ includes the latest observation and execution record but excludes the verification judgment being generated. The Verify Agent produces qualitative feedback $p_{t+1}$ that answers $q_k$, identifies supporting observational evidence, and reports unmet or uncertain conditions. It is instructed to assess completion against the observed outcome, with planned actions and expected landmarks defining the conditions to check. The feedback is then appended to the recent interaction history. For example, in Figure~\ref{fig:framework}, a question requiring both the green sofa and the doorway to be visible after a turn remains unsatisfied if only the sofa is observed.

% Role: verification feedback as a coordination signal.
For an incomplete goal, the Verify Agent provides feedback on what remains to be achieved, guiding the Visuomotor Agent's next interaction. A mismatch between the goal and the scene supports goal revision, while verified completion supports advancement and triggers memory compression. For the final goal, the Verify Agent assesses the instructed stopping condition and informs the stop decision. These judgments are based on the observations available during navigation.

\subsection{Visuomotor Agent: Goal-Conditioned Execution}
\label{sec:execution}

% Role: action selection under the current goal and progress feedback.
The Visuomotor Agent translates the active goal into navigation actions using the current observation, historical context, and verification feedback:
\begin{equation}
    a_t=\mathcal{E}(x,g_k,q_k,o_t,c_t,p_t),
    \label{eq:execute}
\end{equation}
where $p_t$ is the latest feedback for the active goal and is empty when no such judgment is available. The action interface supports forward movement, left and right turns, and stopping. A single interaction may execute a short sequence of movements before returning to visual assessment.

% Role: returning execution evidence to the collaborative loop.
The resulting observation and execution record provide evidence for the next verification step and subsequent goal updates. The local goal can thus remain stable across several interactions while actions adapt to new observations and progress feedback.

\subsection{Memory Agent: Progress-Aware Multimodal Compression}
\label{sec:compression}

% Role: compression boundary, memory update, and retained information.
To reduce repeated processing of completed interactions, the Memory Agent compresses historical context at \emph{verified goal-completion boundaries}. Let $b$ denote a step at which goal completion is verified, and let $H_b$ contain the multimodal interaction history accumulated since the previous compression boundary. The agent updates the existing memory $m^{-}$ as
\begin{equation}
    m^{+}=\mathcal{M}(x,m^{-},H_b,g_k,q_k,p_b),
    \label{eq:summary}
\end{equation}
where $g_k$ is the completed goal, $p_b$ is its verification feedback, and $m^{+}$ is the updated memory. The update consolidates the completed stage with earlier navigation history into three components: a summary of previously executed goals that records the traversed route, encountered landmarks, and their spatial relations; the associated verification outcomes; and a visual keyframe from each verification point. Together, these components retain a compact record of navigation progress and its observational evidence to support subsequent goal generation, execution, and verification.

% Role: context replacement and preservation of unresolved history.
After compression, the updated memory replaces the detailed completed segment in the shared context. Subsequent decisions use this memory alongside the original instruction, current observation, and recent interactions. Evidence for the active goal and unfinished task state remains available. 
% Role: pointer to the detailed execution protocol.
Appendix~\ref{app:prompts} formalizes the agent state transitions, verification-driven routing, and history replacement, with the complete interaction loop in Algorithm~\ref{alg:navharness}.

%% file: sections/experiments_revised_v2.tex
% Spacing changes are local to Experiments.
\begingroup
\setlength{\textfloatsep}{8pt plus 2pt minus 2pt}
\setlength{\floatsep}{6pt plus 2pt minus 2pt}
\setlength{\intextsep}{6pt plus 2pt minus 2pt}
\setlength{\abovecaptionskip}{4pt}
\setlength{\belowcaptionskip}{0pt}
\setlength{\columnsep}{12pt}
\section{Experiments}
\label{sec:experiments}

\subsection{Experimental Setup}
\paragraph{Benchmarks and interface.}
We evaluate 100 episodes each on R2R-CE and RxR-CE~\citep{krantz2020vlnce,ku2020rxr}. R2R-CE uses the Open-Nav subset shared by prior agents~\citep{qiao2025opennav,shi2025smartway,li2026agenticnav}; RxR-CE generally involves longer routes and more detailed instructions. Codex CLI and \method share backbones, episodes, and observation/action interfaces: both navigate from instructions and forward-facing RGB observations using forward movements, turns, and STOP, without reference trajectories or evaluator feedback.

\paragraph{Metrics.}
We report navigation error (NE, meters), oracle success rate (OSR), success rate (SR), success weighted by path length (SPL), and normalized dynamic time warping (nDTW). These assess final proximity, entry into the success region, successful termination, path efficiency, and reference-route agreement, respectively. Lower NE is better; other metrics are percentages with higher values preferred. Context consumption complements these measures to assess the cost of retaining navigation history.

\subsection{Navigation Performance}
\label{sec:nav-results}
\input{tables/navigation}
\paragraph{Comparison settings.}
Table~\ref{tab:navigation} compares \method with Codex CLI under matched GPT-5.6-Sol and GPT-6-Astra backbones to assess adaptive goals and completion verification. While Memory compression is disabled (\textit{Off}) in Table~\ref{tab:efficiency}.

\paragraph{Results and analysis.}
With GPT-6-Astra, \method achieves 79.0\% and 82.5\% SR on R2R-CE and RxR-CE, improving over Codex CLI by 5.0 and 6.2 points. SPL increases by 4.1 and 4.7 points, while NE decreases by 1.22\,m and 0.72\,m. GPT-5.6-Sol likewise gains 4.4 and 6.0 SR points; on RxR-CE, NE falls from 9.03\,m to 7.28\,m and nDTW rises from 41.2 to 47.0. These improvements, particularly the larger SR gains on RxR-CE, support coupling local execution to observed progress: adaptive goals reflect the encountered scene, while verification grounds advancement in completed outcomes rather than issued actions. Using monocular input in a zero-shot setting, GPT-6-Astra-based \method exceeds Qwen-RobotNav by 6.9 and 6.0 SR points. It leads the reported SR on both benchmarks and NE and nDTW on RxR-CE.

\subsection{Navigation Quality and Context Efficiency}
\label{sec:efficiency}
\begin{table}[!t]
\input{tables/efficiency}
\end{table}
% Start beside the accounting paragraph so the figure fits on this page.
\begin{wrapfigure}{r}{0.55\textwidth}
\centering
\includegraphics[width=\linewidth]{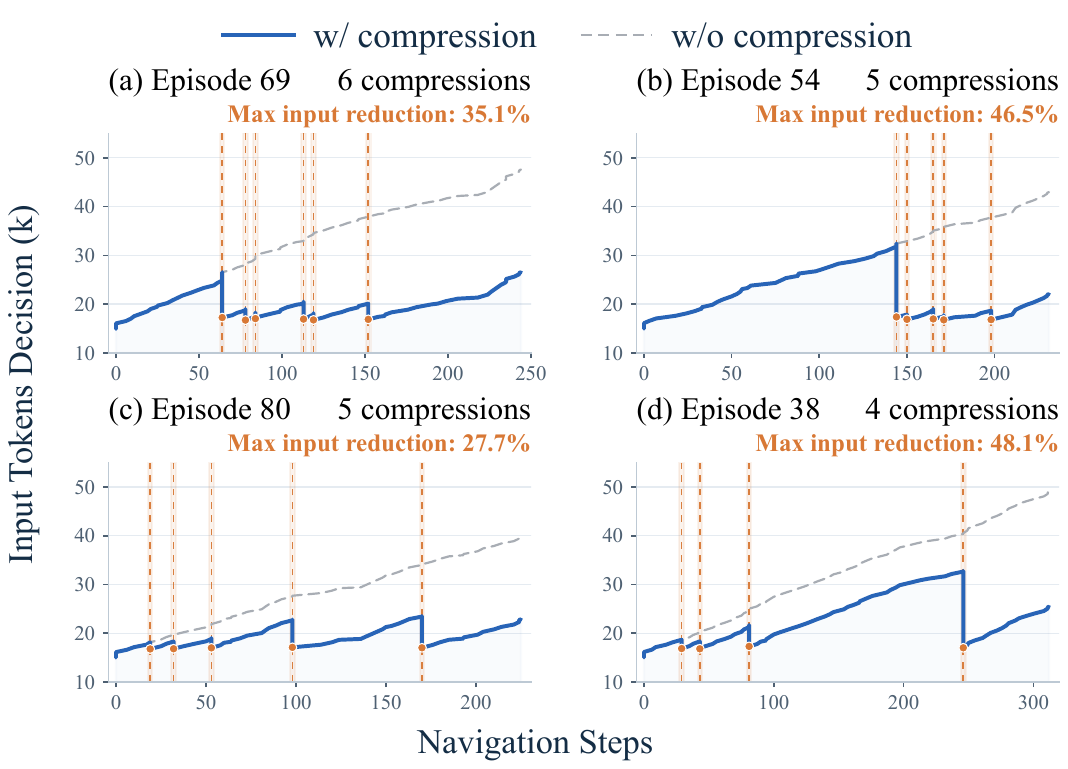}
\caption{\textbf{Input context during navigation.} Vertical markers indicate compression events; percentages report selected local input reductions, not episode-level savings.}
\label{fig:input-four-cases}
\end{wrapfigure}

\paragraph{Comparison settings and accounting.}
Table~\ref{tab:efficiency} compares Codex CLI's native context management (\textit{Native}) with \method without compression (\textit{Off}) and with verified-completion compression (\textit{Goal}) on both benchmarks and backbones. Off versus Goal tests the Memory Agent's addition. Input and Total report mean episode-level tokens ($10^3$) across all components, including compression; Input includes cached tokens, and Total adds output. Image/Dec. and Text/Dec. report input images and text length (k characters) per external decision, potentially comprising multiple actions. Total ratio is $100\,T_{\mathrm{Goal}}/T_{\mathrm{Off}}$; efficiency means exclude incomplete usage records (Appendix~\ref{app:accounting}). Figure~\ref{fig:input-four-cases} complements these episode-level measures with decision-input traces for GPT-6-Astra on four RxR-CE episodes.

\paragraph{Results and analysis.}
Goal reduces total tokens in all four settings. GPT-5.6-Sol saves 35.94\% on R2R-CE and 56.59\% on RxR-CE, with SR decreases of 1 and 2 points relative to Off; R2R-CE SPL improves from 47.9 to 50.3. GPT-6-Astra saves 5.52\% and 4.05\%, with SR decreases of 2 and 1.5 points and an RxR-CE SPL decline from 64.7 to 58.3, however, compressed SR remains above Native throughout. Text/Dec. falls by 55.5--71.6\%, showing that completed-history summaries substantially reduce repeated textual input. Figure~\ref{fig:input-four-cases} reveals the corresponding temporal behavior: compressed traces repeatedly drop at goal-completion boundaries, whereas uncompressed traces continue to grow. The four episodes contain several compressions, with annotated local reductions of 27.7--48.1\%. Verified completion thus provides recurring boundaries for consolidating history while retaining context for ongoing navigation. It is worth noting that local reductions differ from cumulative savings, which include internal calls, compression overhead, and subsequent execution. Progress-aware memory therefore reduces decision context, but the overall trade-off depends on the backbone: GPT-5.6-Sol gains substantial savings, whereas GPT-6-Astra shows modest savings and a larger RxR-CE path-efficiency penalty.

\subsection{Framework Ablation Across Backbones}
\label{sec:framework-ablation}
\begin{wrapfigure}{r}{0.50\textwidth}
\centering
\includegraphics[width=\linewidth]{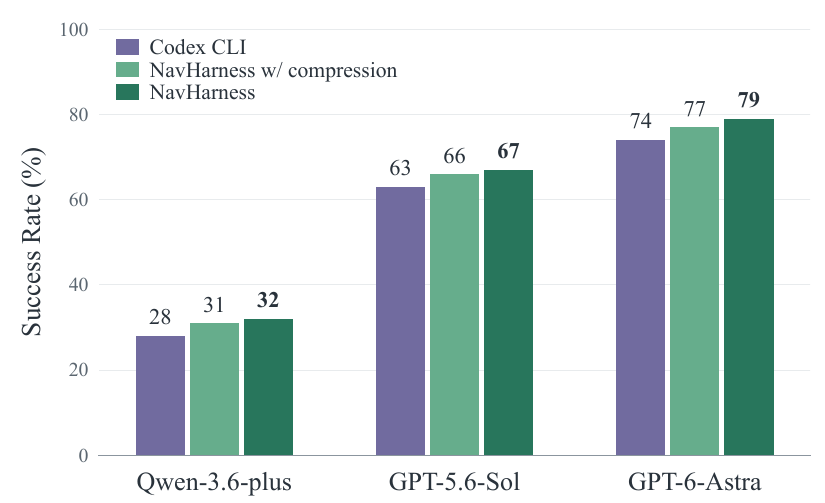}
\caption{\textbf{Framework ablation.} R2R-CE SR (\%), rounded. Plain \method is Off.}
\label{fig:model-comparison}
\end{wrapfigure}
Figure~\ref{fig:model-comparison} compares Codex CLI with \method, with and without compression, on R2R-CE using Qwen-3.6-plus, GPT-5.6-Sol, and GPT-6-Astra. This tests whether the framework benefits backbones with different baseline capabilities.

Without compression, SR increases from 28\%, 63\%, and 74\% to 32\%, 67\%, and 79\%, respectively. With compression, SR remains higher at 31\%, 66\%, and 77\%. The gains across all three backbones indicate that the benefit of goal-guided, verified execution is not confined to the strongest model. Compression preserves part of this advantage. This framework-level comparison does not isolate goal generation from verification.

\subsection{Real-World Navigation}
\label{sec:real}
\begin{wraptable}{r}{0.44\textwidth}
\begingroup
\let\footnotesize\scriptsize
\input{tables/real_world}

\endgroup
\end{wraptable}
We deploy \method on a Unitree Go2 quadruped using only RGB observations from a front-mounted Intel RealSense D435i (Appendix~\ref{app:real}). Eight routes with reference lengths of approximately 15--20\,m span corridors, sofa areas, classrooms, stairs, laboratories, and outdoor spaces, requiring ordered landmark decisions and scene transitions. Each route is tested three times (24 trials). Table~\ref{tab:real} reports SR and mean NE alongside recorded baseline aggregates.

\method achieves 83.3\% SR and 1.51\,m NE, improving over JanusVLN, the strongest listed baseline, by 33.3 SR points and 0.83\,m NE. These results support the real-world effectiveness of \method. The following cases illustrate how observed scene changes during actions inform the assessment of instruction-specific goal completion.

\paragraph{Indoor scene transitions.}
The indoor route in Figure~\ref{fig:real-world-main} follows corridor and lounge landmarks to a yellow chair with a white ball. The robot passes the black chair, aligns with the doorway, enters the room, and approaches the target. Crucially, clearing the doorway precedes the right turn, and seeing the chair precedes the final approach. These distinct stages illustrate how goal-specific spatial conditions connect local execution to completion assessment, rather than treating landmark visibility as sufficient progress.

\begin{figure}[!htb]
\centering
\includegraphics[width=\textwidth]{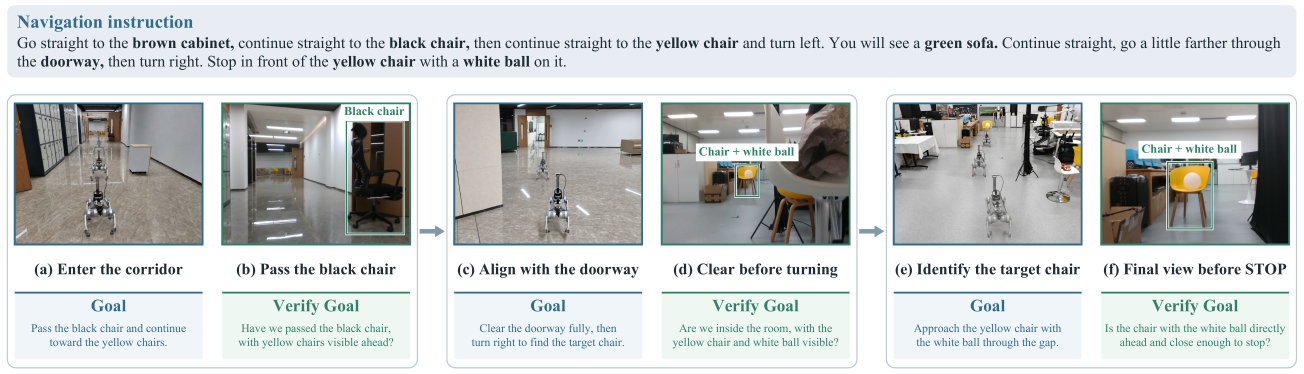}
\caption{\textbf{Indoor scene transitions.} Following ordered landmarks, clearing the doorway before turning, and approaching the yellow chair with a white ball.}
\label{fig:real-world-main}
\end{figure}

\paragraph{Adaptive outdoor navigation.}
The two routes in Figure~\ref{fig:outdoor-case02} test adaptation and ordered progress. The sculpture instruction requires passing the bushes and building corridor before approaching the sculpture; the bicycle instruction requires reaching the green bicycle before turning left onto the ramp, then turning right through the corridor toward the trees. The latter explicitly distinguishes arriving beside a landmark from merely seeing it at a distance.

In the sculpture sequence, the robot realigns with the destination after an obstacle-induced adjustment and resumes its approach. In the bicycle sequence, it approaches the landmark before turning, reaches the upper landing, and continues toward the trees. These behaviors illustrate complementary roles of adaptive goals and verification: the immediate maneuver can change without replacing the global destination, whereas advancement depends on the spatial condition attached to the active goal. Appendix~\ref{app:real-cases} provides further cases.
% Allow the outdoor cases to use the remaining space on the current page.
\setlength{\intextsep}{4pt plus 2pt minus 2pt}
\begin{figure}[!htbp]
\centering
\includegraphics[width=\textwidth]{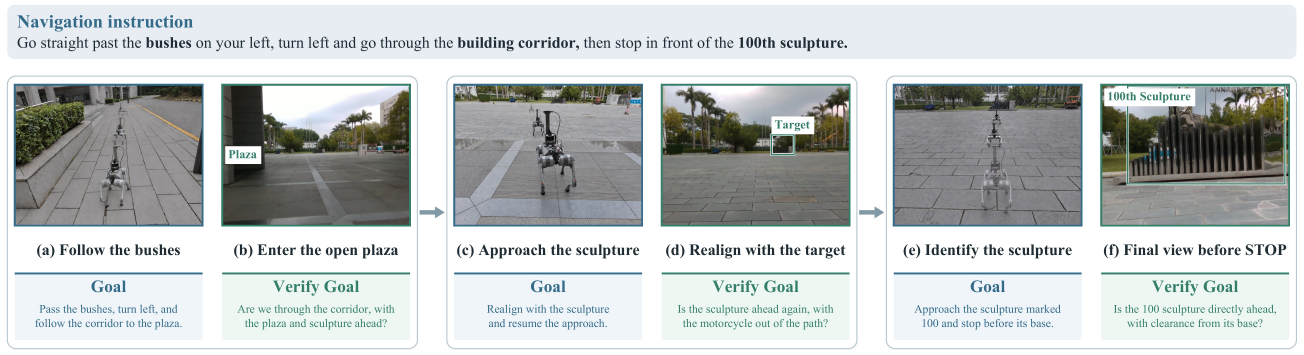}\par
\vspace{4pt}
\includegraphics[width=\textwidth]{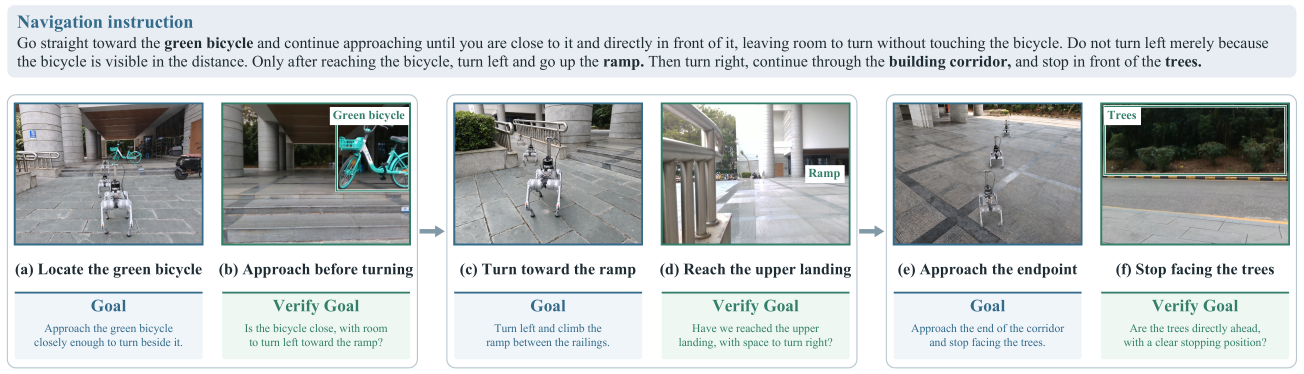}
\caption{\textbf{Adaptive outdoor navigation.} Obstacle-responsive goal adaptation toward the sculpture (top) and landmark-conditioned progress toward the trees (bottom).}
\label{fig:outdoor-case02}
\end{figure}

% The lower sequence combines an initial run with its resumed continuation; its endpoint view alone does not establish a verified successful stop.

\FloatBarrier
\endgroup

%% file: tables/navigation.tex
% Preamble requirement: \usepackage{colortbl} (xcolor is already loaded).
\begin{table}[t]
\centering
\caption{Navigation performance on R2R-CE and RxR-CE. \textbf{Bold}: best reported value per metric; \underline{underline}: best baseline within each group unless bold. Shaded rows denote NavHarness. M = monocular, P = panorama, D = depth; 3-cam = three fixed RGB cameras. NE is in meters; other metrics are percentages. Results retain their original protocols: SmartWay and AgenticNav use Open-Nav\textquotesingle s 100 episodes; Dashes denote unreported or pending results.}
\label{tab:navigation}
\begingroup
\definecolor{NavHarnessTint}{RGB}{232,244,238}
\scriptsize
\setlength{\tabcolsep}{2.2pt}
\renewcommand{\arraystretch}{1.05}
\begin{tabular}{@{}p{155pt}ll*{8}{c}@{}}
\toprule
& & & \multicolumn{4}{c}{R2R-CE} & \multicolumn{4}{c}{RxR-CE} \\
\cmidrule(lr){4-7}\cmidrule(l){8-11}
Method / backbone & Views & Training & NE\down & OSR\up & SR\up & SPL\up & NE\down & SR\up & SPL\up & nDTW\up \\
\specialrule{\lightrulewidth}{\aboverulesep}{0.4pt}
\multicolumn{11}{@{}c@{}}{\textit{Trained navigation models}}\\
\specialrule{\lightrulewidth}{0.4pt}{\belowrulesep}
NaVid~\citep{zhang2024navid} & M & Trained & 5.47 & 49.1 & 37.4 & 35.9 & 8.41 & 23.8 & 21.2 & \pending \\
NaVILA~\citep{cheng2024navila} & M & Trained & 5.22 & 62.5 & 54.0 & 49.0 & 6.77 & 49.3 & 44.0 & 58.8 \\
StreamVLN~\citep{wei2025streamvln} & M & Trained & 4.90 & 63.6 & 56.4 & 50.2 & 5.65 & 54.4 & 45.4 & 63.7 \\
JanusVLN~\citep{zeng2026janus} & M & Trained & 4.78 & 65.2 & 60.5 & 56.8 & 6.06 & 56.2 & 47.5 & 62.1 \\
NavFoM~\citep{zhang2026navfom} & P & Trained & 4.61 & 72.1 & 61.7 & 55.3 & 4.74 & 64.4 & 56.2 & 65.8 \\
AwareVLN~\citep{guo2026awarevln} & M & Trained & 4.02 & 73.5 & 65.4 & 55.1 & 3.95 & 67.6 & 56.1 & 65.7 \\
OmniNav~\citep{xue2026omninav} & M & Trained & 3.74 & 74.6 & 69.5 & 66.1 & 3.77 & 73.6 & 62.0 & \pending \\
Qwen-RobotNav-8B~\citep{qwen2026robotnav} & P & Trained & \underline{3.53} & \underline{78.5} & \underline{72.1} & \underline{66.6} & \underline{3.58} & \underline{76.5} & \textbf{65.7} & \underline{72.5} \\
\specialrule{\lightrulewidth}{\aboverulesep}{0.4pt}
\multicolumn{11}{@{}c@{}}{\textit{Navigation workflows}}\\
\specialrule{\lightrulewidth}{0.4pt}{\belowrulesep}
SmartWay / GPT-4o~\citep{shi2025smartway} & P+D & Zero-shot & 7.01 & 51.0 & 29.0 & 22.46 & \pending & \pending & \pending & \pending \\
SmartWay / GPT-5.5~\citep{shi2025smartway} & P+D & Zero-shot & 5.16 & 60.0 & 44.0 & 35.04 & \pending & \pending & \pending & \pending \\
Vesta~\citep{bjorck2026vesta} & M & Trained & 5.16 & 61.4 & 55.5 & 50.8 & \pending & \pending & \pending & \pending \\
InternVLA-N1 / DualVLN~\citep{wei2025dualvln} & M & Trained & 4.05 & 70.7 & 64.3 & 58.5 & 4.58 & 61.4 & 51.8 & \underline{70.0} \\
ABot-N1~\citep{gong2026abotn1} & 3-cam & Trained & \underline{3.32} & \underline{75.2} & \underline{70.9} & \textbf{67.5} & \underline{3.13} & \underline{73.9} & \underline{63.9} & \pending \\
\specialrule{\lightrulewidth}{\aboverulesep}{0.4pt}
\multicolumn{11}{@{}c@{}}{\textit{Agentic VLN systems}}\\
\specialrule{\lightrulewidth}{0.4pt}{\belowrulesep}
AgenticNav / GPT-5.5~\citep{li2026agenticnav} & P+D & Zero-shot & 5.19 & 65.0 & 55.0 & 48.41 & \pending & \pending & \pending & \pending \\
% Minimal / fable-5-max~\citep{zhou2026mip} & M & Zero-shot & \underline{3.84} & \textbf{83.0} & \underline{78.0} & \underline{65.27} & \pending & \pending & \pending & \pending \\
HarnessVLN / GPT-5.5~\citep{chen2026harnessvln} & P+D & Zero-shot & 4.01 & 72.7 & 60.8 & 43.5 & 6.42 & 53.9 & 38.0 & 54.8 \\
\addlinespace[1pt]
Codex CLI / GPT-5.6-Sol & M & Zero-shot & 4.75 & 70.0 & 62.6 & 44.5 & 9.03 & 28.0 & 21.0 & 41.2 \\
Codex CLI / GPT-6-Astra & M & Zero-shot & 4.49 & 75.0 & 74.0 & 62.5 & \underline{3.14} & \underline{76.3} & \underline{60.0} & \underline{71.9} \\
\specialrule{\lightrulewidth}{2pt}{0pt}
\rowcolor{NavHarnessTint}
\textbf{NavHarness / GPT-5.6-Sol} & M & Zero-shot & 4.60 & 69.5 & 67.0 & 47.9 & 7.28 & 34.0 & 25.0 & 47.0 \\
\rowcolor{NavHarnessTint}
\textbf{NavHarness / GPT-6-Astra} & M & Zero-shot & \textbf{3.27} & \textbf{81.5} & \textbf{79.0} & \underline{66.6} & \textbf{2.42} & \textbf{82.5} & \underline{64.7} & \textbf{73.8} \\
\specialrule{\heavyrulewidth}{0pt}{0pt}
\end{tabular}
\endgroup
\end{table}

%% file: tables/efficiency.tex
\begingroup
\definecolor{NavHarnessTint}{RGB}{232,244,238}
\centering
\setlength{\abovecaptionskip}{0pt}
\setlength{\belowcaptionskip}{4pt}
\caption{\textbf{Navigation quality and context consumption.} Cache/Input/Total: $10^3$ tokens per episode. Cache is included in Input. Image/Dec.: mean input images per external decision. Text/Dec.: mean input text length ($10^3$ characters) per external decision. Total ratio: Goal/Off (\%).}
\label{tab:efficiency}
\scriptsize
\setlength{\tabcolsep}{3pt}
\renewcommand{\arraystretch}{1.05}
\begin{tabular}{@{}llc*{8}{r}@{}}
\toprule
\multicolumn{3}{c}{} & \multicolumn{2}{c}{\makebox[0pt]{Navigation Quality}} & \multicolumn{3}{c}{Token Consumption} & \multicolumn{2}{c}{Decision Context} & Compression \\
\cmidrule(lr){4-5}\cmidrule(lr){6-8}\cmidrule(lr){9-10}\cmidrule(l){11-11}
Model & Harness & Compress. & SR\up & SPL\up & Cache & Input\down & Total\down & Image/Dec.\down & Text/Dec. (k)\down & Total ratio (\%)\down \\
\midrule
\multicolumn{11}{c}{\textit{R2R-CE}} \\
\midrule
GPT-5.6-Sol & Codex CLI & Native & 62.6 & 44.5 & 1351.40 & 1398.72 & 1410.31 & 12.43 & 68.99 & \textemdash \\
 & NavHarness & Off & 67.0 & 47.9 & 1234.50 & 1334.64 & 1345.23 & 13.10 & 80.57 & \textemdash \\
\rowcolor{NavHarnessTint}
 & NavHarness & Goal & 66.0 & 50.3 & 770.45 & \textbf{851.85} & \textbf{861.82} & \textbf{11.80} & \textbf{32.31} & \textbf{64.06\%} \\
\addlinespace[3pt]
GPT-6-Astra & Codex CLI & Native & 74.0 & 62.5 & 656.30 & 687.77 & 689.81 & 9.42 & 73.72 & \textemdash \\
 & NavHarness & Off & 79.0 & 66.6 & 510.55 & 577.91 & 580.91 & 11.28 & 81.15 & \textemdash \\
\rowcolor{NavHarnessTint}
 & NavHarness & Goal & 77.0 & 65.6 & 478.34 & \textbf{544.82} & \textbf{548.87} & \textbf{7.21} & \textbf{ 36.12} & \textbf{94.48\%} \\
\midrule
\multicolumn{11}{c}{\textit{RxR-CE}} \\
\midrule
GPT-5.6-Sol & Codex CLI & Native & 28.0 & 21.0 & 1946.31 & 2017.16 & 2027.81 & 34.59 & 78.65 & \textemdash \\
 & NavHarness & Off & 34.0 & 25.0 & 4008.33 & 4277.80 & 4298.55 & 45.05 & 122.09 & \textemdash \\
\rowcolor{NavHarnessTint}
 & NavHarness & Goal & 32.0 & 21.9 & 1702.70 & \textbf{1847.59} & \textbf{1865.92} & \textbf{18.14} & \textbf{34.65} & \textbf{43.41\%} \\
\addlinespace[3pt]
GPT-6-Astra & Codex CLI & Native & 76.3 & 60.0 & 1097.85 & 1145.12 & 1149.50 & 26.67 & 81.61 & \textemdash \\
 & NavHarness & Off & 82.5 & 64.7 & 1392.89 & 1533.90 & 1538.84 & 27.26 & 105.92 & \textemdash \\
\rowcolor{NavHarnessTint}
 & NavHarness & Goal & 81.0 & 58.3 & 1326.07 & \textbf{1467.08} & \textbf{1476.45} & \textbf{13.62} & \textbf{38.50} & \textbf{95.95\%} \\
\bottomrule
\end{tabular}\par
\endgroup

%% file: tables/real_world.tex
\centering
\captionof{table}{\textbf{Real-world navigation.} Baselines use the supplied aggregates; \method uses eight routes with three trials each.}
\label{tab:real}
\footnotesize
\begin{tabular*}{\linewidth}{@{\extracolsep{\fill}}lrr@{}}
\toprule
Method & SR (\%)\up & NE (m)\down \\
\midrule
NaVILA & 20.83 & 3.91 \\
AwareVLN & 20.83 & 3.89 \\
StreamVLN & 25.00 & 3.78 \\
JanusVLN & 50.00 & 2.34 \\
\midrule
\textbf{NavHarness} & \textbf{83.30} & \textbf{1.51} \\
\bottomrule
\end{tabular*}

%% file: sections/conclusion.tex
\section{Limitations and Conclusion}
\label{sec:conclusion}
\method organizes vision-language navigation around adaptive goals that connect action, verification, and context management. Our central proposal is to use observed task progress both to guide the next objective and to identify completed history that can be summarized. This organization retains the global instruction while allowing execution to respond to the encountered environment.

The current deployment has two practical limitations. First, reliance on remotely served closed-source models incurs substantial inference costs and communication latency, slowing the navigation loop. Second, a single RGB camera provides limited visual coverage: obstacles outside its field of view can remain unobserved, leading to unsafe movements of the quadruped robot. Future work will investigate lightweight VLMs for local deployment to reduce model-serving costs and communication delays, and explore navigation with depth sensing or panoramic observations to improve spatial awareness and obstacle coverage.

%% file: sections/appendix.tex
\section{Extended Related Work}
\label{app:related}

\paragraph{Navigation policies and foundation models.}
Learned navigation policies connect instruction grounding, historical observations, and action selection. Recurrent VLN-BERT maintains a cross-modal state~\citep{hong2021vlnbert}, HAMT explicitly encodes observation and action history~\citep{chen2021hamt}, and DUET combines local grounding with global topological planning~\citep{chen2022duet}. In continuous environments, waypoint prediction bridges high-level decisions and movement~\citep{hong2022bridging}, while ETPNav integrates online topological planning with obstacle-avoiding control~\citep{an2025etpnav}. Scaling instruction--trajectory data further improves learned navigation and generalization~\citep{wang2023scalevln}. More recent navigation models use video-based action prediction~\citep{zhang2024navid}, language-mediated locomotion~\citep{cheng2024navila}, and navigation-oriented reasoning representations~\citep{zhou2024navgpt2}. Qwen-RobotNav extends this line toward a scalable policy with an agent-facing interface~\citep{qwen2026robotnav}. These works establish the value of learned representations and action priors. \method studies how an inference-time framework organizes a general-purpose model's navigation goals and execution history.

\paragraph{Language-model agents for embodied interaction.}
Embodied language-model systems connect semantic reasoning to executable actions and environmental feedback. SayCan grounds skill selection in learned affordances~\citep{ichter2023saycan}; Inner Monologue incorporates scene descriptions and success feedback into planning~\citep{huang2023innermonologue}. ReAct interleaves reasoning with actions, and Reflexion retains verbal feedback for subsequent trials~\citep{yao2023react,shinn2023reflexion}. In VLN, NavGPT explores explicit reasoning from textual observations~\citep{zhou2024navgpt}, DiscussNav organizes multi-expert discussions~\citep{long2024discussnav}, and MapGPT uses a map to guide adaptive planning~\citep{chen2024mapgpt}. Open-Nav studies zero-shot navigation with open-source language models in continuous environments~\citep{qiao2025opennav}. Recent tool-calling agents and harnesses give the model greater authority over interaction and coordinate memory, feedback, and recovery~\citep{li2026agenticnav,zhou2026mip,chen2026harnessvln,amap2026abot}. This literature motivates both the reasoning model and the organization of its execution. \method focuses on a shared goal representation through which execution, progress assessment, and context management remain connected.

\paragraph{Progress monitoring and corrective navigation.}
Explicit progress assessment has long been part of instruction-following navigation. The Self-Monitoring agent learns auxiliary progress estimates from visual--textual grounding~\citep{ma2019selfmonitoring}; the Regretful Agent uses progress estimates to support backtracking~\citep{ma2019regretful}. More recently, AwareVLN learns structured reasoning about spatial state and task progress~\citep{guo2026awarevln}, while SmartWay combines waypoint prediction with backtracking for zero-shot navigation~\citep{shi2025smartway}. These methods show that progress estimation and recovery are established research questions. Our verification questions make the completion condition explicit for each goal generated from the encountered situation. The resulting judgment informs goal continuation, revision, and advancement, and verified completion also defines when the corresponding history can be compressed. Our focus is this coupling of adaptive goals, observable completion, and context boundaries.

\paragraph{Context management for navigation.}
Navigation memory must retain useful evidence while controlling the cost of processing past interaction. StreamVLN uses slow--fast context modeling and cache reuse~\citep{wei2025streamvln}, while JanusVLN separates semantic and spatial information into dual implicit memories~\citep{zeng2026janus}. Language-based navigation can also summarize history externally, as in NavGPT~\citep{zhou2024navgpt}. In general language-model settings, LLMLingua-2 learns task-agnostic extractive prompt compression~\citep{pan2024llmlingua2}; studies of long-context question answering and retrieval show that supplying more context does not ensure effective use of relevant information~\citep{liu2024lostmiddle}. Those text-task findings motivate careful context design without establishing a navigation failure mechanism. \method uses verified goal completion as a navigation-specific boundary for summarizing completed interaction. It preserves information required by the remaining instruction and keeps recent evidence for the active goal. The distinction concerns when to compress and what navigation state to retain; summarization itself is an established technique.

\section{Evaluation Protocols}
\label{app:protocols}

\paragraph{Observation and action interface.}
The simulation interface provides a forward-facing RGB observation and four primitive actions: move forward by $0.25$\,m, turn left by $15^{\circ}$, turn right by $15^{\circ}$, and STOP. An external decision may request a short sequence of primitives. Execution returns an observation and a record of the actions performed, which are supplied to the Verify Agent for progress assessment. STOP terminates the episode.

\paragraph{Online information.}
The instruction and observations are the task evidence supplied to the model. Historical records distinguish actions, observations, goal descriptions, verification judgments, and interpretations. The model does not receive the evaluator's reference path, success label, or distance to the destination. The exact image selection policy determines which archived observations are actually attached to a request; a textual mention of a historical frame does not imply that the frame is visible to the model.

\paragraph{Comparison settings.}
We compare Codex CLI and \method using matched backbones, episode lists, and observation/action interfaces. Table~\ref{tab:navigation} evaluates \method with memory compression disabled to assess the combined effect of adaptive goals and completion verification. Table~\ref{tab:efficiency} additionally compares this configuration with goal-based memory compression enabled. Codex CLI retains its native context management.

\paragraph{External references.}
Published methods can differ in training data, visual input, learned executors, and the set of evaluated episodes. The table therefore separates published context from the controlled comparison. A full validation split, a fixed subset, and a subset of a previously released pool are different cohorts even when they share the same benchmark name.

\paragraph{Published configurations in Table~\ref{tab:navigation}.}
The trained-policy rows use the main configurations reported by their authors. StreamVLN uses the RGB-only model with additional VLN training data in the revised paper's Table~I, without voxel pruning. NavFoM uses four RGB views, and Qwen-RobotNav uses the panoramic 8B policy without its agentic planner. NaVid's RxR results are cross-dataset transfer results from a model trained on R2R; they are not RxR-trained results. ABot-N1 uses front, left, and right cameras, represented as 3-cam rather than a full panorama. The method groups distinguish trained navigation policies, navigation workflows, and agentic VLN systems according to their execution organization, following \citet{zhou2026mip}.

SmartWay/GPT-4o is the original configuration; SmartWay/GPT-5.5 is the explicit reproduction reported by \citet{li2026agenticnav}. Both and AgenticNav use the Open-Nav 100-episode R2R-CE subset. HarnessVLN's main results follow the val-unseen protocol in its Table~2; the fixed subsets specified for its ablations are not assumed to define its main evaluation. Zero-shot identifies the reasoning model's navigation-task adaptation status; learned navigation tools may still be present. These external results provide context rather than a single matched ranking.

\section{Agent Formulation, Coordination, and Prompt Templates}
\label{app:prompts}
This section specifies the state transitions and coordination rules underlying the four agents, followed by their prompt templates. Algorithm~\ref{alg:navharness} provides the interaction loop using the same $\mathcal{G}$, $\mathcal{V}$, $\mathcal{E}$, and $\mathcal{M}$ as Section~\ref{sec:method}. The formalization makes explicit when history is appended, when a goal persists or changes, and which verified events permit memory replacement. The prompt templates summarize the instructions provided to each agent and are edited for readability. Algorithm~\ref{alg:navharness} specifies their invocation order and the updates to the shared navigation state.

\begin{algorithm}[!htbp]
\caption{NavHarness: Goal-Verified Navigation}
\label{alg:navharness}
\fontsize{8.5}{9.2}\selectfont
\algrenewcommand{\algorithmicindent}{0.8em}
\renewcommand{\algorithmiccomment}[1]{\unskip\hspace{0.4em}$\triangleright$~\textcolor{gray}{#1}}
\begin{algorithmic}
\Require Instruction $x$, initial observation $o_0$, interaction budget $T$
\Ensure Executed navigation trajectory $\mathcal{T}$
\end{algorithmic}
\noindent\begin{minipage}[t]{0.54\linewidth}
\begin{algorithmic}[1]
\State $t\gets 0$, $k\gets 0$; $m,r,p_0\gets\emptyset$; $\mathcal{T}\gets(o_0)$
\State $(g_k,q_k)\gets\mathcal{G}(x,o_0,(m,r))$ \Comment{Goal and verification question}
\While{$t<T$ and the episode is active}
    \State $a_t\gets\mathcal{E}(x,g_k,q_k,o_t,(m,r),p_t)$ \Comment{Move, turn, or stop}
    \State Execute $a_t$; receive $o_{t+1}$ and execution record $e_{t+1}$
    \State Append $(g_k,q_k,o_t,a_t,e_{t+1},o_{t+1})$ to $r$
    \State $p_{t+1}\gets\mathcal{V}(x,g_k,q_k,o_{t+1},(m,r))$ \Comment{Verify observed progress}
    \State Append $p_{t+1}$ to $r$ and $(a_t^{\mathrm{exec}},o_{t+1})$ to $\mathcal{T}$
    \State $t\gets t+1$
    \If{$p_t$ verifies completion of $g_k$}
        \State $m\gets\mathcal{M}(x,m,r,g_k,q_k,p_t)$ \Comment{Retain relevant state and keyframes}
        \State $r\gets\emptyset$ \Comment{Replace the summarized segment}
    \EndIf
\algstore{navharnessloop}
\end{algorithmic}
\end{minipage}\hfill
\begin{minipage}[t]{0.43\linewidth}
\begin{algorithmic}[1]
\algrestore{navharnessloop}
    \If{the episode has terminated}
        \State \textbf{break}
    \EndIf
    \If{$p_t$ verifies the instructed stopping condition}
        \State Execute \textsc{Stop} through $\mathcal{E}$; append it to $\mathcal{T}$; \textbf{break}
    \EndIf
    \If{$p_t$ verifies goal completion or indicates that the goal needs revision}
        \State $k\gets k+1$; $(g_k,q_k)\gets\mathcal{G}(x,o_t,(m,r))$
        \State $p_t\gets\emptyset$ \Comment{Earlier feedback remains in context}
    \EndIf
\EndWhile
\State \Return $\mathcal{T}$
\end{algorithmic}
\end{minipage}
\end{algorithm}

\FloatBarrier

\Needspace{10\baselineskip}
\subsection{Initializing the Navigation Episode}
The initialization prompt establishes the route instruction, observation and action interface, and movement budget. The first observation supplies the initial goal and its verification question before movement begins. Initialization is performed once per episode; resuming an ongoing episode preserves the active goal and supplied observation.
\paragraph{State and update order.}
The interaction state separates persistent memory from the unsummarized records of the current navigation segment. Let $k_t$ be the value of the goal index $k$ at the start of interaction $t$ in Algorithm~\ref{alg:navharness}. We write
\begin{equation}
\begin{aligned}
z_t&=(o_t,m_t,r_t,g_{k_t},q_{k_t},p_t),
&c_t&=(m_t,r_t),\\
z_0&=(o_0,\emptyset,\emptyset,g_0,q_0,\emptyset),
&\mathcal{T}_0&=(o_0).
\end{aligned}
\label{eq:app-state}
\end{equation}
Here $(g_0,q_0)$ is generated from the initial observation as in Equation~\ref{eq:goal}. The instruction $x$ remains fixed and available to every agent. A complete interaction first records execution, then obtains verification feedback, then updates memory, and finally decides whether to stop, replace the goal, or continue. This ordering determines which version of the history each agent can access. In the equations below, $\widehat p_{t+1}$ denotes the newly returned verification judgment before a possible reset; Algorithm~\ref{alg:navharness} stores this judgment in $p_{t+1}$ and later clears it if the goal changes.

\begin{navprompt}{Initialization Prompt}
You are controlling a robot using its forward-facing RGB camera. Follow this navigation instruction to its endpoint:

[Navigation instruction]

Available actions are STOP, move forward 0.25 m, turn left 15 degrees, and turn right 15 degrees. You have [movement budget] actions. STOP permanently ends the episode.

At episode start, obtain the first RGB observation, initial adaptive goal, and verification question. If navigation has already started, continue from the current goal and supplied view without repeating initialization.

After moving, use the resulting camera view returned with the execution record. Work autonomously until you stop.
\end{navprompt}
The instruction remains available throughout execution. The initial observation starts the shared navigation context; subsequent observations, verification feedback, and memory updates populate that context as the episode proceeds. The real-robot briefing identifies the physical platform and adds motion-uncertainty and clearance guidance rather than assuming exact simulator motion.

\Needspace{10\baselineskip}
\subsection{Generating the Next Goal}
Goal generation receives the complete route instruction, the current observation, and available execution history. Historical and current images are distinguished, and records identify which actions were executed and which statements were interpretations. 
\paragraph{Goal persistence and replacement.}
Goal generation is conditional on the coordination decision rather than repeated at every interaction. Let $u_{t+1}$ be the routing decision defined in Equation~\ref{eq:app-routing}, and let $\eta_{t+1}=\mathbf{1}[u_{t+1}=\textsc{UpdateGoal}]$. The goal state evolves as
\begin{equation}
\begin{aligned}
k_{t+1}&=k_t+\eta_{t+1},\\
(g_{k_{t+1}},q_{k_{t+1}})&=
\begin{cases}
\mathcal{G}(x,o_{t+1},(m_{t+1},r_{t+1})),&\eta_{t+1}=1,\\
(g_{k_t},q_{k_t}),&\eta_{t+1}=0,
\end{cases}\\
p_{t+1}&=
\begin{cases}
\emptyset,&\eta_{t+1}=1,\\
\widehat p_{t+1},&\eta_{t+1}=0.
\end{cases}
\end{aligned}
\label{eq:app-goal-lifecycle}
\end{equation}
The memory and recent-history arguments on the right are the \emph{post-update} values from Equation~\ref{eq:app-memory-transition}, matching the order in Algorithm~\ref{alg:navharness}. Thus, advancing after completion uses the newly consolidated memory, whereas revising an incomplete goal retains its detailed execution evidence. Resetting the active feedback prevents a judgment about the previous goal from being applied to its replacement; that judgment remains available in the recorded or summarized history.

\begin{navprompt}{Goal Agent Prompt}
Infer the current route progress from the instruction, observations, and execution history.

Generate only the next goal that can be pursued from the present situation; the goal may require several actions.

Pair it with a question that tests both completion and the expected landmark or spatial relation.

Treat expected observations as conditions to check, not as established facts.

For the final goal, check arrival without requiring that the stop command has already been issued.
\end{navprompt}
The returned goal and question are kept together. This pairing makes the completion condition available before the actor begins the next stage and allows later verification to refer to the same intended outcome.

\Needspace{10\baselineskip}
\subsection{Checking Observed Completion}
Verification receives the original instruction, active goal, paired question, and available observations and execution history. 
\paragraph{Evidence before and after verification.}
Let $\oplus$ denote ordered concatenation of records, and define the execution event
$v_t=(g_{k_t},q_{k_t},o_t,a_t,e_{t+1},o_{t+1})$.
The two history states surrounding a verification call are
\begin{equation}
\begin{aligned}
r_{t+1}^{-}&=r_t\oplus\langle v_t\rangle,
&c_{t+1}^{-}&=(m_t,r_{t+1}^{-}),\\
r_{t+1}^{+}&=r_{t+1}^{-}\oplus\langle\widehat p_{t+1}\rangle.
\end{aligned}
\label{eq:app-verification-history}
\end{equation}
Equation~\ref{eq:verify} consumes $c_{t+1}^{-}$: it includes the executed actions and resulting observation, but not the judgment being generated. The judgment is appended only afterward. This ordering separates the evidence available to the Verify Agent from the judgment it produces.

\paragraph{Feedback-driven routing.}
Write $d_{t+1}$ for verified completion of the active goal, $\rho_{t+1}$ for a judgment that the goal needs revision, and $\sigma_{t+1}$ for verification of the instructed stopping condition. These Boolean indicators denote the corresponding conditions in Algorithm~\ref{alg:navharness}; they are interpretations of $\widehat p_{t+1}$, not additional classifiers or numerical confidence thresholds. Let $\chi_{t+1}$ indicate that the environment has already terminated. After the memory update, coordination follows
\begin{equation}
u_{t+1}=
\begin{cases}
\textsc{Halt},&\chi_{t+1}=1,\\
\textsc{Stop},&\chi_{t+1}=0,\ \sigma_{t+1}=1,\\
\textsc{UpdateGoal},&\chi_{t+1}=0,\ \sigma_{t+1}=0,\ d_{t+1}\lor\rho_{t+1},\\
\textsc{Continue},&\text{otherwise}.
\end{cases}
\label{eq:app-routing}
\end{equation}
The cases make the algorithm's priority explicit: termination and the instructed stopping condition take precedence over generating another goal. Completion and revision can both lead to a goal update, but only completion triggers compression. Uncertainty that does not support either condition leaves the goal active and carries the feedback into the next interaction.

\begin{navprompt}{Verify Agent Prompt}
Answer the verification question from recorded execution and visual evidence.

Check the actions actually performed and whether the observed result matches the expected outcome.

Base completion on the latest observation; intended actions and expected scenes alone do not prove success.

Explain the evidence, completion or uncertainty, and the appropriate next step.
\end{navprompt}
The feedback records the observed evidence, the goal's completion status, and a recommendation for the next interaction. It is appended to the shared history and used to determine whether to continue execution, revise the goal, or advance after verified completion.

\Needspace{10\baselineskip}
\subsection{Executing the Active Goal}
The Visuomotor Agent receives the original instruction, active goal and verification question, latest RGB observation, and available navigation history, including memory and verification feedback. Its prompt connects these inputs to environment actions while allowing one goal to persist across multiple interactions.
\paragraph{Batched actions and realized execution.}
One external decision can contain several primitive commands. With
$\mathcal{A}=\{\textsc{Forward},\textsc{Left},\textsc{Right},\textsc{Stop}\}$,
the selected batch and its realized prefix are
\begin{equation}
\begin{aligned}
a_t&=(\alpha_{t,1},\ldots,\alpha_{t,L_t})\in\mathcal{A}^{+},\\
a_t^{\mathrm{exec}}&=(\alpha_{t,1},\ldots,\alpha_{t,J_t}),
\qquad 0\leq J_t\leq L_t,\\
\alpha_{t,j}&\neq\textsc{Stop}\qquad\text{for }1\leq j<J_t.
\end{aligned}
\label{eq:app-action-prefix}
\end{equation}
The environment may end a batch before every requested command is executed, and no later command is executed after STOP. The record $e_{t+1}$ and resulting observation therefore determine what occurred; the proposed batch alone does not establish its outcome. The trajectory in Algorithm~\ref{alg:navharness} records the executed action prefix $a_t^{\mathrm{exec}}$ and the resulting observation. The interaction history additionally retains the requested batch and execution record:
\begin{equation}
\begin{aligned}
\mathcal{T}_{t+1}&=\mathcal{T}_t\oplus\langle a_t,o_{t+1}\rangle,\\
N_{t+1}^{\mathrm{move}}&=N_t^{\mathrm{move}}+
\sum_{j=1}^{J_t}\mathbf{1}[\alpha_{t,j}\neq\textsc{Stop}].
\end{aligned}
\label{eq:app-execution-accounting}
\end{equation}
The counter starts at $N_0^{\mathrm{move}}=0$ and increases with each executed movement command. The environment enforces the primitive-action budget and terminates the episode when that budget is exhausted. This budget is distinct from the external-interaction bound $T$ in Algorithm~\ref{alg:navharness}, since one interaction may execute several primitives. STOP ends the episode without adding a movement to $N^{\mathrm{move}}$.

\begin{navprompt}{Visuomotor Agent Prompt}
Before moving, read the active goal and its verification question. Use the current RGB view, relevant navigation history, and available verification feedback to select actions toward that goal.

Execute the selected movement actions in order, then inspect the returned observation and execution record. To look in another direction, turn and use the resulting view. A goal may span several interactions; do not regenerate it before every action.

After each interaction, submit the returned observation and execution record for verification. Use the feedback to continue toward an incomplete goal, request a corrective goal when the current objective no longer fits the scene, or advance after verified completion.

For an ongoing episode, complete this assessment before selecting the next movement sequence. If an updated goal is already supplied, continue with that goal.

Respect the movement budget. Use STOP to end the episode when you judge that the instructed endpoint has been reached.
\end{navprompt}
Execution supplies the observations and action records used by the other agents. Verification feedback supports the stop decision, but the inspected action interface does not impose an additional mechanical verification gate on STOP. On the physical robot, STOP may also end a run for safety and does not by itself establish successful arrival.

\Needspace{10\baselineskip}
\subsection{Summarizing Completed Navigation}
The summarizer receives the original instruction, previous memory, newly completed execution records, and the attached visual evidence. Its purpose is to preserve information needed for subsequent navigation rather than produce a general description of the scene. 
\paragraph{Completion boundaries and history replacement.}
Let $b_t$ mark the interaction immediately after the most recent completed compression, with $b_0=0$. The segment available at the end of interaction $t$ is
\begin{equation}
H_{t+1}=\bigoplus_{i=b_t}^{t}
\langle v_i,\widehat p_{i+1}\rangle
=r_{t+1}^{+},
\qquad
b_{t+1}=\begin{cases}
t+1,&d_{t+1}=1,\\
b_t,&d_{t+1}=0.
\end{cases}
\label{eq:app-completion-boundary}
\end{equation}
Each $v_i$ retains the goal and question active when that interaction occurred, so a segment may include corrections under several successive goals. The coupled memory/history transition is
\begin{equation}
(m_{t+1},r_{t+1})=
\begin{cases}
\bigl(\mathcal{M}(x,m_t,H_{t+1},g_{k_t},q_{k_t},\widehat p_{t+1}),\emptyset\bigr),
&d_{t+1}=1,\\
(m_t,r_{t+1}^{+}),&d_{t+1}=0.
\end{cases}
\label{eq:app-memory-transition}
\end{equation}
This joint update formalizes the two adjacent operations in Algorithm~\ref{alg:navharness}: generate the replacement memory, then clear the summarized recent segment. The updated memory contains a summary of previously executed goals that records the traversed route, landmarks, and spatial relations; the associated verification outcomes; and an RGB keyframe from each verification point. The textual summary and retained keyframes jointly provide context for subsequent navigation. The original instruction and current observation remain separate inputs. In particular, any interval without verified completion obeys
\begin{equation}
\left[\bigwedge_{i=s}^{t}(d_{i+1}=0)\right]
\quad\Longrightarrow\quad
\begin{cases}
m_{t+1}=m_s,\\
r_{t+1}=r_s\oplus\displaystyle\bigoplus_{i=s}^{t}
\langle v_i,\widehat p_{i+1}\rangle.
\end{cases}
\label{eq:app-revision-preservation}
\end{equation}
This invariant holds even if several goals are revised within the interval: replacement does not certify success or erase unresolved evidence. When completion does trigger compression, unresolved conditions needed for later navigation must be carried into the replacement memory rather than treated as completed merely because the raw segment is cleared.

\begin{navprompt}{Memory Agent Prompt}
Combine the previous memory with the new execution records and observations.

Summarize the previously executed goals as a record of the traversed route, relevant landmarks, and spatial relations. Preserve the associated verification outcomes, including corrections and uncertainties relevant to the remaining instruction.

Distinguish observations and executed actions from interpretations, and preserve uncertainty.

Describe the latest goal's execution separately from earlier relevant navigation history without duplicating the same information.
\end{navprompt}
Verified goal completion triggers memory compression. Revising an incomplete goal preserves the unsummarized interaction history so that subsequent decisions retain the evidence needed for recovery.

\section{Metric Definitions and Context Accounting}
\label{app:accounting}

\paragraph{Navigation metrics.}
We use the benchmark evaluators to compute navigation metrics.
Navigation error (NE) measures the final distance to the goal.
Success rate (SR) measures successful termination, whereas oracle
success rate (OSR) measures whether the trajectory reaches the
success region at any point. Success weighted by path length (SPL)
accounts for both success and path efficiency, and normalized dynamic
time warping (nDTW) measures agreement with the reference route.
NE is reported in meters, with lower values preferred; the remaining
metrics are percentages, with higher values preferred.
The real-world evaluation reports SR and final NE under the protocol
described in Appendix~\ref{app:real}.

\paragraph{Episode-level token consumption.}
For episode $e$, let $\mathcal{Q}_e$ denote the model calls with
recorded usage across goal generation, verification, execution,
and memory compression, including recorded retries. We aggregate
input, cached input, and total tokens as
\begin{equation}
    I_e = \sum_{j\in\mathcal{Q}_e} I_j,\qquad
    C_e = \sum_{j\in\mathcal{Q}_e} C_j,\qquad
    T_e = \sum_{j\in\mathcal{Q}_e}(I_j+O_j),
    \label{eq:tokens}
\end{equation}
where $I_j$, $C_j$, and $O_j$ denote input, cached input, and output
tokens for call $j$, respectively.
Table~\ref{tab:efficiency} reports their means per episode in
thousands of tokens. Efficiency averages include successful and
failed episodes with complete usage records and exclude incomplete
records.

Cached tokens are included in Input and are not added again when
computing Total. When a provider reports cached and uncached input
separately, the two fields are combined once. Reasoning tokens already
included in the reported output are likewise not counted twice.
The Cache column describes input reuse rather than context removal
and therefore has no preferred direction.

\paragraph{Inputs per external decision.}
An external decision is one logical request to the Visuomotor Agent
and may produce multiple environment actions.
Image/Dec.\ measures the mean number of input image instances per
external decision, while Text/Dec.\ measures the mean input text
length in thousands of characters.
These input counts include repeated history and inputs supplied to
the internal agents; internal goal, verification, and memory calls
do not add external decisions to the denominator.
Text/Dec.\ measures characters rather than tokens and therefore
cannot be directly substituted for Input or Total.
Likewise, image counts do not specify visual token consumption,
which depends on the backbone and its visual encoding.

\paragraph{Total token ratio.}
Table~\ref{tab:efficiency} distinguishes three configurations:
\textit{Native} uses Codex CLI's native context management;
\textit{Off} uses \method with memory compression disabled; and
\textit{Goal} enables compression at verified goal-completion
boundaries.
For each backbone and benchmark, the reported Total ratio compares
the mean episode-level consumption of Goal and Off:
\begin{equation}
    R_{\mathrm{total}}
    =
    100\,
    \frac{\overline{T}_{\mathrm{Goal}}}
         {\overline{T}_{\mathrm{Off}}},
    \label{eq:total-ratio}
\end{equation}
where $\overline{T}$ denotes the mean total tokens per episode.
The ratio is reported as a percentage; values below $100\%$
indicate lower consumption with compression.
The corresponding percentage saving is $100-R_{\mathrm{total}}$.
Both configurations include all recorded inference components,
and Goal includes the Memory Agent's compression overhead.
We interpret these savings together with SR and SPL because
compression can also change the executed trajectory, number of
model calls, and termination time.

\paragraph{Local context evolution.}
Figure~\ref{fig:input-four-cases} plots the recorded input tokens
for external navigation decisions against navigation steps.
Vertical markers indicate compression events, and the annotations
report relative input decreases at selected events.
These values describe local changes in the observed input sequence;
they are not an aggregate reduction over all compression events.
The compressed and uncompressed runs can follow different
trajectories, so equal navigation-step indices do not necessarily
correspond to identical observations or decision contexts.

\section{Real-World Evaluation Protocol}
\label{app:real}
We evaluate eight challenging routes with reference lengths of approximately 15--20\,m across corridors, sofa areas, classrooms, stairs, laboratories, and outdoor spaces. Each route is tested three times with \method, yielding 24 trials. The platform is a Unitree Go2 (approximately $70\times40$\,cm in length and width) with a front-mounted Intel RealSense D435i camera approximately 70\,cm above the ground. The USB-connected camera supplies $640\times480$ RGB images at a configured 30\,FPS; depth, infrared, and camera IMU streams are disabled. Communication uses ROS~2 Foxy with CycloneDDS. Table~\ref{tab:real} reports aggregate SR and mean NE across routes, retaining the recorded baseline values.

\section{Additional Real-World Cases}
\label{app:real-cases}
The following examples complement the indoor and outdoor cases in Section~\ref{sec:real}. Each figure presents six egocentric observations selected to illustrate route transitions, local corrections, and the final approach. Goal and verification annotations are scene-aligned explanatory reconstructions, rather than verbatim logged outputs; colored outlines are editorial annotations. The sequences show observed behavior and do not independently establish the causal contribution of an individual framework component.

\begin{figure}[p]
\centering
\setlength{\abovecaptionskip}{2pt}
\setlength{\belowcaptionskip}{0pt}

\includegraphics[
    width=\linewidth,
    height=0.17\textheight,
    keepaspectratio
]{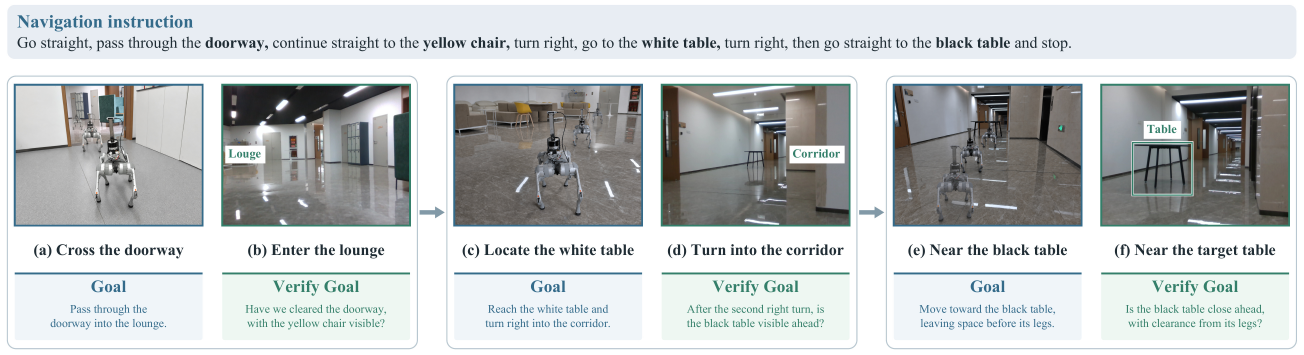}
\caption{\textbf{Successive landmark turns.} The route links a doorway, a yellow chair, a white table, and a black table through successive landmark-conditioned turns. Panel (h) shows a clear view during the final approach; the run subsequently calls STOP.}
\label{fig:real-world-case01}

\vspace{4pt}

\includegraphics[
    width=\linewidth,
    height=0.17\textheight,
    keepaspectratio
]{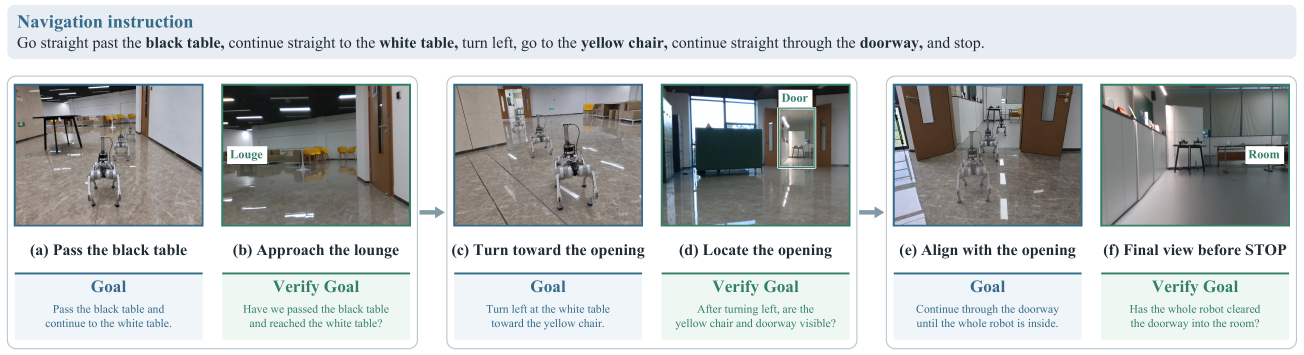}
\caption{\textbf{Landmark following and doorway clearance.} After following the table landmarks, the agent approaches the doorway and makes local corrections before entering the room. Panel (h) is the final observation before the recorded stop action.}
\label{fig:real-world-case02}

\vspace{4pt}

\includegraphics[
    width=\linewidth,
    height=0.17\textheight,
    keepaspectratio
]{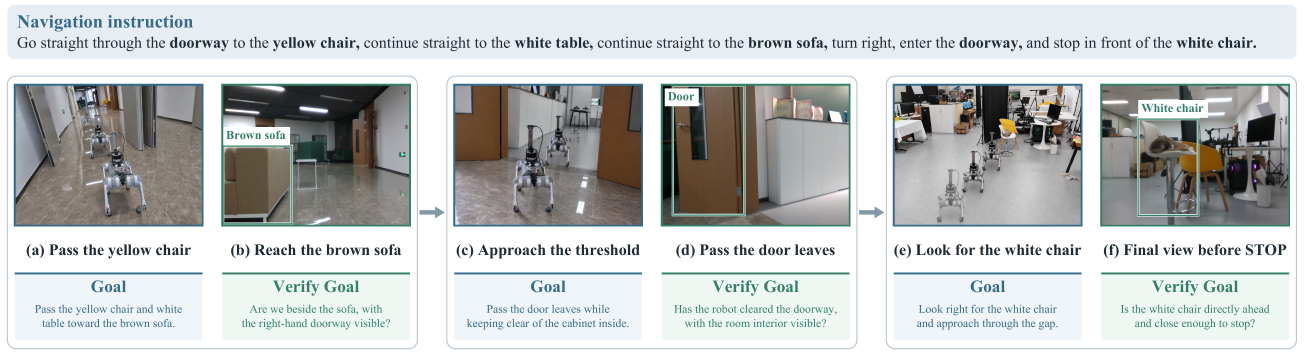}
\caption{\textbf{Narrow doorway and chair identification.} The agent follows lounge landmarks, aligns with a narrow opening formed by the door leaves, and enters the room before locating the instructed white chair. Panel (h) is the final observation before the recorded stop action.}
\label{fig:real-world-case03}

\vspace{4pt}

\includegraphics[
    width=\linewidth,
    height=0.17\textheight,
    keepaspectratio
]{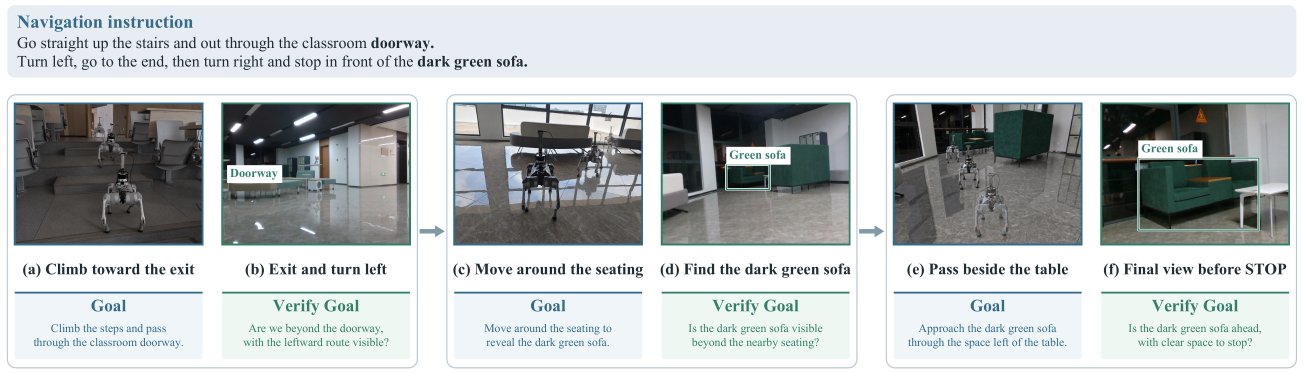}
\caption{\textbf{Stepped exit and sofa approach.} The route begins at a stepped doorway and continues through a furnished area. The local objective adapts to the seating arrangement before the final approach to the dark green sofa. Panel (h) is the final observation before the recorded stop action.}
\label{fig:real-world-case06}

\end{figure}